\documentclass[runningheads]{llncs}

\usepackage{eccv}
\usepackage{eccvabbrv}

\usepackage{graphicx}
\usepackage{booktabs}
\usepackage{amsmath,amssymb}
\usepackage{multirow}
\usepackage{enumitem}
\usepackage{subcaption}
\usepackage{xspace}
\usepackage{capt-of}
\usepackage{placeins}
\usepackage[accsupp]{axessibility}

\usepackage{hyperref}

\usepackage{orcidlink}

\newcommand{\arc}{ArcFace\xspace}
\newcommand{\ada}{AdaFace\xspace}
\newcommand{\adavit}{AdaFace-ViT\xspace}
\newcommand{\bridged}{{\em Aligned}\xspace}
\newcommand{\native}{{\em Native}\xspace}
\newcommand{\unaligned}{{\em Unaligned}\xspace}

\begin{document}

\title{Unmasking Face Embeddings: Reading, Rendering and Naming with Foundation Models}
\titlerunning{Unmasking Face Embeddings with Foundation Models}

% Anonymized automatically by the `review' option of the eccv package.
\author{Fizza Rubab\orcidlink{0009-0001-6979-7746} \and
Yiying Tong\orcidlink{0000-0002-7929-4333} \and
Arun Ross\orcidlink{0000-0001-8850-3013}}
\authorrunning{F.~Rubab et al.}
\institute{Michigan State University\\
\email{\{rubabfiz, ytong, rossarun\}@msu.edu}}

\maketitle

\vspace{-5.5cm}
\begin{center}
    {\small Paper accepted at ECCV Workshops, September 2026.}
\end{center}
\vspace{4.7cm}

\noindent
\begin{minipage}{\textwidth}
\vspace{-0.575cm}
\centering
\includegraphics[width=0.9\linewidth]{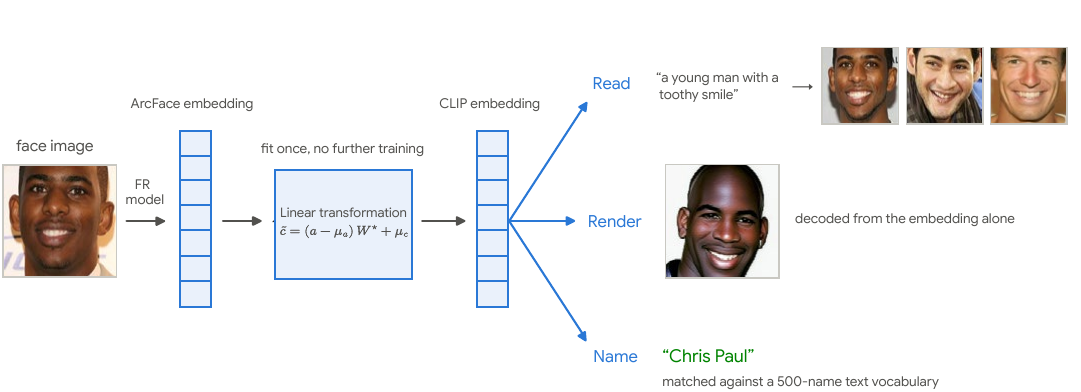}
\vspace{-0.1cm}
\captionof{figure}{\textbf{Overview.} A single linear transformation maps a
face embedding from a face recognition model into a foundation space, where it can be
\emph{read} by a text query, \emph{rendered} by a diffusion decoder, and \emph{named}
against a text vocabulary without additional training.}
\label{fig:overview}
\vspace{-0.3cm}
\end{minipage}

\begin{abstract}
Modern face recognition (FR) owes much of its success to deep neural networks that learn to extract compact identity embeddings from face images. These models are typically trained for identity discrimination, producing embeddings that are highly effective for biometric matching but largely opaque to semantic interpretation. In contrast, foundation models, pretrained on broad visual or vision--language tasks, provide rich interfaces for describing, retrieving, generating, and organizing visual content. This contrast raises a natural question: \textit{what capabilities become available when face embeddings from domain-specific FR models are made interoperable with foundation models?} Building on recent work on embedding compatibility across models, we use simple pre-computed linear transformations, estimated from paired embeddings alone, to connect existing FR models with off-the-shelf foundation models. Once aligned with a foundation model, a face embedding can be ``unmasked'' in multiple ways, without training or modifying either model: it can be \emph{read} in natural language, enabling free-form text queries over a gallery of FR embeddings; \emph{rendered} into a face image that recovers a person's appearance, using an unmodified diffusion decoder; and converted to a {\em name}, enabling identification even in the absence of an enrolled face gallery. In effect, one linear transformation turns an identity embedding into a rich embedding for web-scale foundation models. This interoperability exposes face embeddings as semantically and visually rich biometric representations, with direct implications for interpretability, retrieval, reconstruction, and template security.
\end{abstract}

% =====================================================================
\section{Introduction}
\label{sec:intro}
% =====================================================================
The automated process of recognizing an individual based on their face image is known as face recognition (FR)~\cite{jain2011intro}. FR systems are being used in several biometric applications ranging from unlocking smartphones to identifying perpetrators of crime. Modern FR systems rely on deep neural networks that map a face image to a compact embedding vector~\cite{kim202650}. Vectors corresponding to different face images are compared using distance metrics in order to verify or determine the identity of an unknown face image~\cite{schroff2015facenet}.
Domain-specific models such as ArcFace~\cite{arcface}, CosFace~\cite{wang2018cosface},
and AdaFace~\cite{adaface} are trained using face images and identity labels alone. As a result, their embeddings are effective for biometric matching. However, with no language or semantic supervision, an embedding (i.e., a face template) cannot be directly queried in words, decoded into an image, or associated with a name in the absence of an enrolled face gallery.

In contrast, foundation models, pretrained
on broad heterogeneous data including vision--language data and recently explored for biometric tasks~\cite{sony2025benchmarking,sony2025foundation}, can go beyond image matching. Image--text models such as CLIP~\cite{clip},
MetaCLIP~\cite{metaclip}, and SigLIP~\cite{siglip} enable language-based retrieval of visual content and support semantic description through text. Embedding-conditioned diffusion models~\cite{unclip,kandinsky} invert
an embedding back into an image. Web-scale pretraining even associates face images with
the names of well-known individuals~\cite{idia}.

If FR embeddings could be mapped to the embedding spaces of foundation models, they could avail of the capabilities that these models already provide. However, such a mapping is not
guaranteed: various FR and foundation models are trained using different data, objectives, loss functions, architectures, and embedding
dimensions. Yet a growing body of work shows that independently trained networks learn
geometrically related representations (embeddings)~\cite{relreps,platonic,lenc,bansal}. For face
images specifically, recent work has shown that FR and foundation-model embeddings are
linearly compatible~\cite{compat}. \textit{So the question now is: how can aligning embeddings across FR and foundation models be further exploited?}

Our approach is deliberately minimal (Fig.~\ref{fig:overview}). A single linear
transformation, estimated once from paired embeddings between two models, aligns a domain-specific FR model to an off-the-shelf foundation model. The transformation uses no attribute
labels, fine-tunes neither model, and retrains no decoder, text encoder, or downstream
head. The transformed face embedding can be utilized directly by the foundation model. We view this as an \emph{unmasking} mechanism:
alignment exposes semantic, visual, and identity-related information that is
present in an FR embedding but inaccessible through the original model.\footnote{Code: \url{https://github.com/Fizza-Rubab/unmasking-face-embeddings}}

We study three such implications of this alignment. An embedding can be \textit{read}: free-form text queries
can search an aligned gallery of FR embeddings, retrieving the faces that match the description (Sec.~\ref{sec:language}). It can be
\textit{rendered}: an unmodified diffusion decoder can reconstruct the person's appearance
and soft-biometric attributes, though not fine identity details (Sec.~\ref{sec:generation}).
Further, it can be \textit{named}: the aligned embedding can be used to find out a person's name, identifying well-known individuals from the embedding alone
(Sec.~\ref{sec:naming}). The performance of the aligned embeddings on these tasks is compared against the foundation model's native performance, as well as against two baselines: unaligned embeddings and randomly transformed embeddings (Sec.~\ref{sec:method}).
This ensures that every capability we report is attributable to the learned alignment
rather than other incidental factors.

Our main contributions are:
\begin{enumerate}[noitemsep,topsep=2pt]
\item Demonstrating that a single pre-computed linear transformation
makes a gallery of face embeddings searchable using natural language text, at close to the foundation
model's own retrieval accuracy (Sec.~\ref{sec:language}).
\item Showing that a linear transformation into a diffusion decoder's conditioning space inverts an FR embedding into a realistic face image that preserves appearance and soft-biometric attributes (Sec.~\ref{sec:generation}).
\item Matching an aligned embedding against text embeddings of
candidate names, thereby identifying well-known individuals from their face embeddings even in the absence of an enrolled face gallery
(Sec.~\ref{sec:naming}).
\end{enumerate}

% =====================================================================
\section{Related Work}
\label{sec:related}
% =====================================================================
\noindent\textbf{Face and Foundation Representations.}
Face-specific models are optimized primarily for identity discrimination, from
FaceNet~\cite{schroff2015facenet} and CosFace~\cite{wang2018cosface} to
ArcFace~\cite{arcface}, AdaFace~\cite{adaface}, MagFace~\cite{meng2021magface}, and the
keypoint-guided KPRPE~\cite{kprpe}. Foundation models are large-scale models pretrained on broad and diverse data to learn general-purpose representations that can support a wide range of downstream tasks. Contrastive image--text models~\cite{clip,metaclip,siglip}, for example, align visual and textual features, while diffusion decoders~\cite{unclip,kandinsky} built on latent diffusion~\cite{ldm} can map learned image representations back into images. We use publicly available pretrained models from both families without fine-tuning (Table~\ref{tab:models}), fitting only a linear
transformation between their embedding spaces.

\noindent\textbf{Representation Similarity and Alignment.}
A long line of work shows that independently trained networks learn geometrically related
representations. Lenc and Vedaldi~\cite{lenc} swapped layers of different networks joined
by a linear transform, and Bansal \etal~\cite{bansal} revisited model stitching as a way
to compare representations. Relative representations~\cite{relreps} align latent spaces
through shared anchors, and the Platonic representation hypothesis~\cite{platonic} frames
cross-model convergence as an emergent property of scale. For face images in particular, Rubab \etal~\cite{compat} found that independently trained FR models, and even FR
and general foundation models, occupy linearly compatible spaces, so that a low-capacity
linear transformation improves cross-model recognition performance from chance to competitive accuracy. While~\cite{compat} establishes that model spaces can be aligned, in this work we study what the alignment enables.

\noindent\textbf{Reading and Reconstructing Face Embeddings.}
Several lines of work extract more than a match score from a face embedding.
Terh{\"o}rst \etal~\cite{terhorst2020beyond} showed that FR embeddings retain
soft-biometric information well beyond identity, by training supervised classifiers on
the embeddings themselves. Retrieving faces from language is typically pursued
by training a dedicated cross-modal model on paired face--caption data, as in the
FaceCaption-15M dataset and its FLIP model~\cite{facecaption}. Face embeddings are also
partially invertible: reconstruction networks~\cite{mai2019reconstruction} and
diffusion-based inverters~\cite{dong2023genface} train a dedicated inversion network per
target model, while identity-conditioned generators such as Arc2Face~\cite{arc2face} and
InstantID~\cite{instantid} synthesize faces directly from FR embeddings. Closest to our
setup, Otroshi Shahreza \etal~\cite{shahreza2024adapter} also fit a linear adapter on
paired embeddings, but map into Arc2Face~\cite{arc2face}, a face foundation model trained on 42M faces
whose conditioning space is itself an FR embedding space. The reconstruction fidelity is
inherited from that face-native generator, and the pipeline serves reconstruction alone.
In contrast, we align into generic vision--language spaces with no face-specific
training, using one closed-form linear transformation from which retrieval, reconstruction, and naming
all follow.

\noindent\textbf{Open-Set Recognition and Identity Knowledge in Foundation Models.}
Naming an embedding against a vocabulary of names is an open-set problem, in which a probe
may belong to no gallery class and must be rejected rather than
matched. Naming from an embedding is possible because
foundation models memorize identities: identity-inference attacks show that CLIP
associates names with the faces of individuals in its pretraining data~\cite{idia}.
These attacks probe the foundation model with an \emph{image}; we probe it with an
aligned FR embedding.

% =====================================================================
\section{Method}
\label{sec:method}
% =====================================================================
Our transformation is inspired by the linear embedding compatibility reported in~\cite{compat}. Let $f_{\mathrm{src}}$ be a face model and $f_{\mathrm{dst}}$ a foundation model. For
an image $x$ they produce embeddings $a=f_{\mathrm{src}}(x)$ and $c=f_{\mathrm{dst}}(x)$.
Given a set of paired embeddings $\{(a_i,c_i)\}_{i=1}^{n}$ constituting a training set of images and requiring no labels beyond the pairing, we estimate a linear transformation by least squares:
\begin{equation}
W^\star=\arg\min_{W}\;\sum_{i}\big\lVert (a_i-\mu_a)\,W-(c_i-\mu_c)\big\rVert_2^2,
\qquad
\tilde c=(a-\mu_a)\,W^\star+\mu_c,
\label{eq:transform}
\end{equation}
where, $\mu_a$ and $\mu_c$ are the training-set means of the two spaces. Subtracting the
source mean and adding back the target mean places the transformed embedding $\tilde c$ in the
foundation model's own coordinate frame, which is where its downstream modules operate.
We refer to any such module that inputs the foundation embedding as a \emph{head}. Any head $g$ is then applied to a source
embedding simply as $g(\tilde c)$. Fitting $W^\star$ takes only a few seconds on a CPU and is
done only once per (face, foundation) pair.

\noindent\textbf{Baseline comparison.}
For every task (Secs.~\ref{sec:language}--\ref{sec:naming}) we compare four methods. {\em Native} applies the head to the true
foundation embedding $c$ and gives a ceiling. {\em Aligned} applies the head to $\tilde c$ and
is our method. {\em Unaligned} applies the head to the raw source embedding itself, zero-padded or
truncated to the target dimension and scaled to the target norm; it tests whether the two
spaces happen to be usable without any alignment. \emph{Random} applies the head to an
embedding transformed by a random matrix of the same shape as $W^\star$, and controls for
everything except the learned weights. The consistent gap between these two baselines ({\em Unaligned} and \emph{Random})
and our {\em Aligned} method is the effect we attribute to compatibility.

\noindent\textbf{Matching the head's input space.}
Different heads expect embeddings from different encoders. For example, the two diffusion decoders used in this paper are conditioned on the image embeddings of two different CLIP encoders: ViT-bigG for Kandinsky and ViT-H for Stable unCLIP (Sec.~\ref{sec:setup}), and not on the CLIP ViT-B/32 space used for text retrieval. The training images are embedded with the encoder required by that head, and these embeddings serve as the targets $c_i$ in Eq.~\eqref{eq:transform}. The transformed FR embedding then lies in the space that the head expects.

\noindent\textbf{Scoring against text.}
For tasks involving text embeddings, namely retrieval and naming, we center both
visual and text embeddings before comparing: probe embeddings by the training-set mean of aligned embeddings and
text anchors by the mean of the text vocabulary. This removes the average offset that
separates the image and text sub-spaces of contrastive models. For name determination, each text anchor is averaged over four fixed prompt templates
(Sec.~\ref{sec:naming}). Both the centering means and the prompts are determined from training data only.

\noindent\textbf{Unregularized solution.}
Throughout the
paper, we use the plain, unregularized least-squares solution of Eq.~\eqref{eq:transform}. Using ridge regularization only lowered naming accuracy in our ablations, since it suppresses the low-variance source directions where fine identity lives.

% =====================================================================
\begin{table}[tb]
\centering
\caption{Models used in this study, all publicly available and used without fine-tuning.}
\label{tab:models}
\setlength{\tabcolsep}{4pt}
\renewcommand{\arraystretch}{0.9}
\scriptsize
\begin{tabular}{lllcc}
\toprule
& Model & Architecture & Dim & Training data \\
\midrule
\multirow{4}{*}{\shortstack[l]{Face\\(source)}}
 & \arc~\cite{arcface}   & IR-101 CNN & 512 & WebFace4M~\cite{webface260m} \\
 & \ada~\cite{adaface}   & IR-101 CNN & 512 & MS1MV2~\cite{ms1m} \\
 & \adavit~\cite{adaface} & ViT-B     & 512 & WebFace4M~\cite{webface260m} \\
 & KPRPE~\cite{kprpe}    & ViT-B      & 512 & WebFace4M~\cite{webface260m} \\
\midrule
\multirow{5}{*}{\shortstack[l]{Foundation\\(target)}}
 & CLIP~\cite{clip}      & ViT-B/32 & 512 & WIT-400M \\
 & MetaCLIP~\cite{metaclip} & ViT-B/32 & 512 & CC-400M \\
 & SigLIP~\cite{siglip}  & ViT-B/16 & 768 & WebLI \\
 & CLIP (Kandinsky~2.2~\cite{kandinsky}) & ViT-bigG/14 & 1280 & LAION-2B \\
 & CLIP (Stable unCLIP~\cite{unclip})    & ViT-H/14    & 1024 & LAION-2B \\
\bottomrule
\end{tabular}
\end{table}

\section{Experimental Setup}
\label{sec:setup}
% =====================================================================
\noindent\textbf{Models.}
Table~\ref{tab:models} summarizes the FR and foundation models used in this work. Face embeddings are extracted using \arc and \ada
(both IR-101 CNNs), \adavit, a ViT-B trained with the AdaFace objective, and
KPRPE~\cite{kprpe}, the same ViT-B backbone with keypoint-relative position encoding, giving us both CNN and ViT source architectures. For the tasks involving text,
namely retrieval and naming, we use three general-purpose contrastive image--text
models: CLIP~\cite{clip}, MetaCLIP~\cite{metaclip}, which varies the pretraining
corpus, and SigLIP~\cite{siglip}, which varies the contrastive loss. For the generation
task, we use two diffusion decoders, Kandinsky~2.2~\cite{kandinsky} and
Stable unCLIP~\cite{unclip}, which are conditioned on the image embeddings of CLIP
ViT-bigG and CLIP ViT-H, respectively. None of the five foundation targets was trained
with a face-specific objective and so any capability we recover is borrowed from generic
web-scale pretraining rather than from a face-tuned model. We additionally use DINOv2
ViT-B/14~\cite{dinov2}, which is neither an alignment source nor a target, as an
independent reference space for measuring appearance similarity in the image generation
experiments (Sec.~\ref{sec:generation}).

\noindent\textbf{Datasets and alignment.}
The CFP~\cite{cfp} dataset provides 500 identities with celebrity names and is used for the naming and
generation experiments. UTKFace~\cite{utkface} provides age, gender, and ethnicity labels,
while CelebA~\cite{celeba} provides 40 binary attributes; both are used for the text-based retrieval experiments.
Each transformation is fit on a 70\% training split and evaluated on the remaining
30\%. Splits are identity-disjoint wherever identity labels exist and so no test
identity is seen during alignment. On CFP, this yields 3{,}500 training and 1{,}500 test pairs from 350 and 150 disjoint
identities, respectively. On CelebA, it yields 28{,}031 training and 11{,}969 test
pairs from 6{,}556 and 2{,}810 disjoint identities. UTKFace lacks identity labels and
is split by image: 16{,}593 training and 7{,}112 test pairs. Since this split is not
identity-disjoint, we verified the retrieval results by re-clustering UTKFace faces by FR
similarity and repeating the evaluation on an identity-disjoint split. mAP changed by
at most $0.03$ (e.g., CLIP \native changed from $0.768$ to $0.741$), leaving the conclusions unchanged.
All reported numbers in the experimental results are on the held-out test data.

% =====================================================================
\section{Searching Embeddings with Language}
\label{sec:language}
% =====================================================================
In a typical biometric system, an enrolled face gallery can normally be searched only with another face image. We make
it searchable with text. Every embedding in the face gallery is aligned into a foundation model's
joint image--text space, and a natural-language query is encoded by that model's text
encoder and compared against the aligned gallery using cosine similarity.

\begin{figure}[tb]
\centering
\includegraphics[width=0.82\linewidth]{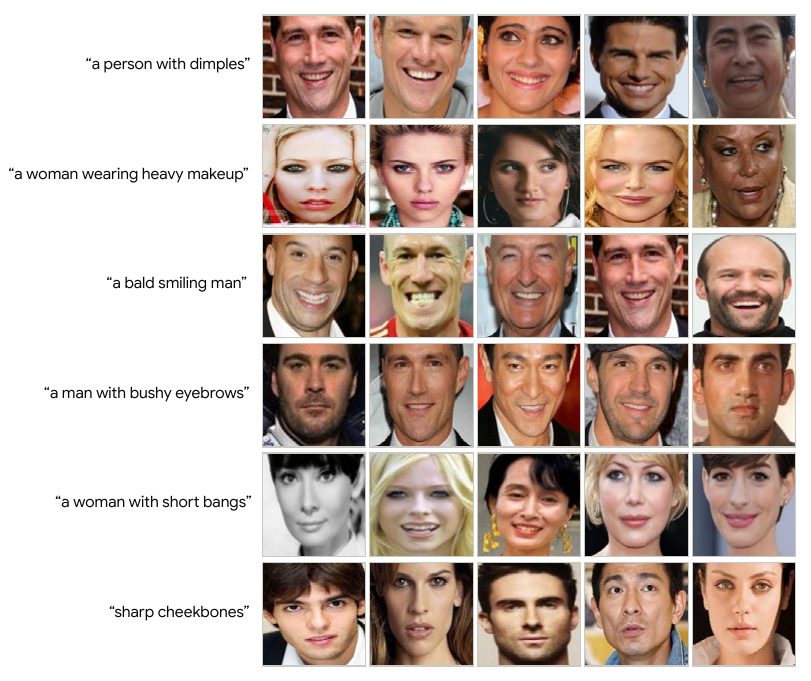}
\caption{Free-form retrieval: top-5 (one image per identity) for each prompt, over a
gallery of aligned KPRPE embeddings in CLIP space from held-out CFP identities.}
\label{fig:textmontage}
\end{figure}

We evaluate with two query sets whose relevance is defined by ground-truth labels. On
UTKFace, we use 9 demographic queries derived from its age, gender and ethnicity
annotations (``a photo of a man'', ``an old person'', \ldots). On CelebA, we use 22
queries derived from its 40 binary attributes: 16 single-attribute queries (``a person
wearing eyeglasses'') and 6 compositional sentences whose relevant set is the
conjunction of several attributes (``a smiling young woman with blond hair''). A gallery face is
relevant if and only if its labels satisfy the query, and we report the mean average
precision (mAP): the average precision (AP) of the cosine ranking, averaged over all queries.
Table~\ref{tab:text} reports mAP for all four methods. Aligned retrieval closely
tracks the \native ceiling. On UTKFace, it can exceed the ceiling: aligned \adavit
scores marginally higher than the native embeddings on two of the three targets (0.809 vs.\ 0.795
for MetaCLIP, 0.796 vs.\ 0.762 for SigLIP). A possible reason is that an FR embedding
focuses on the face and discards background context that can mislead the native model.
On the richer CelebA queries, the best aligned source reaches 97--106\% of \native,
again exceeding the ceiling on MetaCLIP. Neither {\em Unaligned} nor \emph{Random} retrieves above chance. The
improvement is therefore attributable to the learned alignment. This pattern holds across all three image--text models, which differ in pretraining
corpus and contrastive objective. Figure~\ref{fig:textmontage} shows top retrievals for free-form prompts. The
retrieved faces match descriptions that the source FR model was never trained to explicitly
represent.

\begin{table}[tb]
\centering
\caption{Text-to-face retrieval: mAP over natural-language queries against a gallery of FR embeddings. Mean\,$\pm$\,std over five
random splits (identity-disjoint on CelebA; by image on UTKFace).}
\label{tab:text}
\setlength{\tabcolsep}{3.2pt}
\footnotesize
\resizebox{0.72\linewidth}{!}{%
\begin{tabular}{llccc}
\toprule
Dataset & Ranker & CLIP & MetaCLIP & SigLIP \\
\midrule
\multirow{7}{*}{UTKFace}
 & \native (ceiling) & 0.767\,{\tiny$\pm$0.002} & 0.795\,{\tiny$\pm$0.002} & 0.762\,{\tiny$\pm$0.003} \\
\cmidrule(lr){2-5}
 & random transform.\ & 0.289\,{\tiny$\pm$0.014} & 0.280\,{\tiny$\pm$0.006} & 0.280\,{\tiny$\pm$0.006} \\
 & \unaligned & 0.278\,{\tiny$\pm$0.001} & 0.297\,{\tiny$\pm$0.001} & 0.282\,{\tiny$\pm$0.001} \\
 & \bridged \arc & 0.733\,{\tiny$\pm$0.004} & 0.780\,{\tiny$\pm$0.003} & 0.783\,{\tiny$\pm$0.002} \\
 & \bridged \ada & 0.722\,{\tiny$\pm$0.004} & 0.773\,{\tiny$\pm$0.004} & 0.779\,{\tiny$\pm$0.002} \\
 & \bridged \adavit & \textbf{0.765}\,{\tiny$\pm$0.004} & \textbf{0.809}\,{\tiny$\pm$0.003} & \textbf{0.796}\,{\tiny$\pm$0.003} \\
 & \bridged KPRPE & 0.759\,{\tiny$\pm$0.003} & 0.806\,{\tiny$\pm$0.004} & 0.794\,{\tiny$\pm$0.002} \\
\midrule
\multirow{7}{*}{CelebA}
 & \native (ceiling) & 0.555\,{\tiny$\pm$0.007} & 0.511\,{\tiny$\pm$0.006} & 0.569\,{\tiny$\pm$0.004} \\
\cmidrule(lr){2-5}
 & random transform.\ & 0.202\,{\tiny$\pm$0.003} & 0.202\,{\tiny$\pm$0.003} & 0.201\,{\tiny$\pm$0.003} \\
 & \unaligned & 0.200\,{\tiny$\pm$0.001} & 0.194\,{\tiny$\pm$0.001} & 0.205\,{\tiny$\pm$0.001} \\
 & \bridged \arc & 0.457\,{\tiny$\pm$0.003} & 0.467\,{\tiny$\pm$0.003} & 0.504\,{\tiny$\pm$0.003} \\
 & \bridged \ada & 0.514\,{\tiny$\pm$0.005} & 0.524\,{\tiny$\pm$0.005} & 0.551\,{\tiny$\pm$0.006} \\
 & \bridged \adavit & \textbf{0.542}\,{\tiny$\pm$0.008} & \textbf{0.542}\,{\tiny$\pm$0.006} & \textbf{0.563}\,{\tiny$\pm$0.006} \\
 & \bridged KPRPE & 0.516\,{\tiny$\pm$0.006} & 0.518\,{\tiny$\pm$0.006} & 0.547\,{\tiny$\pm$0.006} \\
\bottomrule
\end{tabular}}
\end{table}

% =====================================================================
\section{Decoding Embeddings into Faces}
\label{sec:generation}
% =====================================================================
The alignment also provides a generative interface. We transform a face embedding into
the image-embedding space expected by a diffusion decoder and decode it into a face. For each decoder and each method, we evaluate all 1{,}500 held-out CFP test faces along three axes
(Table~\ref{tab:gen}). \emph{Attribute preservation} is the agreement between CLIP
predictions of seven binary attributes (gender, age band, eyeglasses, beard,
expression, hair shade, baldness) on the decoded and the original image. Each attribute is predicted zero-shot, following CLIP-style classification. For a given binary attribute we form two opposing text descriptions, encode each with CLIP's text encoder, and assign the attribute value whose text embedding has the higher cosine similarity with the image embedding. For eyeglasses, for example, the two descriptions are ``a person wearing
eyeglasses'' and ``a person without eyeglasses''.
\emph{Appearance resemblance} is measured by cosine similarity in
DINOv2~\cite{dinov2} space, a model involved nowhere in the pipeline, and by
LPIPS~\cite{lpips}, a learned perceptual distance. \emph{Realism} is measured by
FID~\cite{fid} against the 1{,}500 real CFP test faces. We
also report \emph{identity retention}, the FR cosine similarity, in $[-1,1]$, between
the decoded and original image, as assessed by ArcFace and AdaFace (last two columns
of Table~\ref{tab:gen}).

\begin{figure}[tb]
\centering
\includegraphics[width=0.9\linewidth]{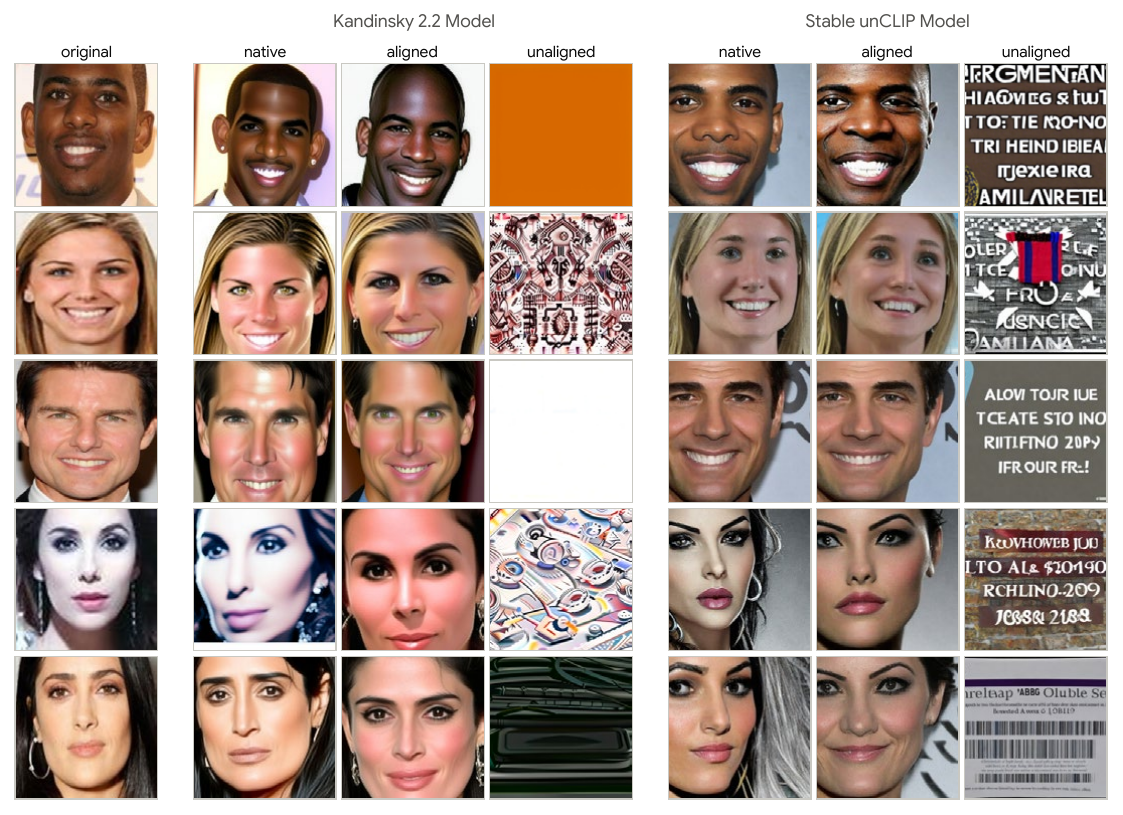}
\caption{Embedding-to-image results for the two decoders: native (ceiling), aligned
KPRPE (ours), and unaligned. The aligned embedding decodes the person's appearance; the
raw unaligned one decodes to non-face imagery.}
\label{fig:gen}
\end{figure}

Table~\ref{tab:gen} summarizes the results. Images recovered from the aligned
embeddings agree with the original on 76--83\% of the attributes, against a \native ceiling of 83--87\%, while {\em Unaligned} and \emph{Random} remain
at 58--63\% (the agreement two unrelated faces exhibit by chance). DINOv2
resemblance shows the same ordering (aligned $\approx$0.53--0.59, native
$\approx$0.68--0.69, random $\approx$0.38--0.42). Aligned reconstructions are realistic: their FID (64--81) lies far below the {\em Unaligned} and \emph{Random} floors (284--326 and 148--190), and is only modestly above the \native ceiling, which attains the best FID of all methods. The two baselines fail in interestingly different ways (Fig.~\ref{fig:gen}): the \unaligned embedding
decodes to non-face imagery (FID $>$280), while the \emph{random}-transformation output collapses to
a near-constant generic face that is plausible as an image but unrelated to the person.

Fine identity does not survive the decoding, and this is a property of the decoder
rather than a failure of the alignment. On CFP, genuine matches score $0.49{\pm}0.17$ under ArcFace and $0.51{\pm}0.17$ under
AdaFace, while impostors score $0.08{\pm}0.12$ under both (mean${\pm}$std over all
within-identity and 60{,}000 random cross-identity pairs). Even the \native method, which inverts
the true foundation embedding, reaches only 0.10--0.17: a CLIP-conditioned decoder
encodes appearance rather than identity. Aligned reconstructions score 0.04--0.16, above \emph{Random} and close to the decoder-limited ceiling; this ordering is consistent under both FR models. We therefore present this ability as appearance recovery, not identity
reconstruction. For reference, a face-native generator (Arc2Face~\cite{arc2face}), conditioned on the original image's ID embedding, recovers more identity (0.24 and 0.28 under ArcFace and AdaFace, Table~\ref{tab:gen}) but still falls well short of a genuine match. This suggests that the identity that survives is limited by the decoder rather than by the alignment alone. Dedicated inversion networks can recover finer identity, but each must be
trained against one specific target model~\cite{mai2019reconstruction,shahreza2024adapter}.
Here nothing is trained: the decoder is untouched, and the only fitted component is the
linear map that also serves retrieval and naming. The privacy implication is nevertheless concrete: a plausible likeness,
with the correct demographics, expression and hairstyle, can be rendered from a stored
template using only publicly available pretrained models and one linear transformation.

\begin{table}[tb]
\centering
\caption{Embedding-to-image on held-out CFP faces. Attr.: agreement of seven
zero-shot CLIP attributes with the original; DINOv2 and LPIPS: appearance; FID: realism;
last two columns: FR-cosine identity retention. Arc2Face is a face-native reference, not
an alignment method.}
\label{tab:gen}
\setlength{\tabcolsep}{4pt}
\footnotesize
\resizebox{\linewidth}{!}{%
\begin{tabular}{llcccc|cc}
\toprule
Decoder & Source & Attr.\ $\uparrow$ & DINOv2 $\uparrow$ & LPIPS $\downarrow$ &
FID $\downarrow$ & \arc cos & \ada cos \\
\midrule
\multirow{7}{*}{\shortstack[l]{Kandinsky 2.2\\(CLIP bigG)}}
 & \native (ceiling) & 0.873 & 0.681 & 0.538 & 58.0 & 0.114 & 0.162 \\
\cmidrule(lr){2-8}
 & random transform. & 0.626 & 0.416 & 0.584 & 148.0 & 0.00 & 0.006 \\
 & \unaligned     & 0.608 & 0.014 & 0.806 & 326.2 & 0.021 & 0.053 \\
 & \bridged \arc  & 0.800 & 0.543 & 0.572 & 66.7 & 0.056 & 0.079 \\
 & \bridged \ada  & 0.804 & 0.557 & 0.573 & 66.8 & 0.045 & 0.098 \\
 & \bridged \adavit & \textbf{0.827} & \textbf{0.593} & \textbf{0.566} & 65.1 & \textbf{0.082} & \textbf{0.114} \\
 & \bridged KPRPE & 0.820 & 0.578 & 0.571 & \textbf{63.8} & 0.077 & 0.107 \\
\midrule
\multirow{7}{*}{\shortstack[l]{Stable unCLIP\\(CLIP ViT-H)}}
 & \native (ceiling) & 0.827 & 0.685 & 0.630 & 57.3 & 0.104 & 0.167 \\
\cmidrule(lr){2-8}
 & random transform. & 0.598 & 0.382 & 0.662 & 189.6 & 0.028 & 0.049 \\
 & \unaligned     & 0.582 & 0.019 & 0.723 & 283.7 & 0.013 & 0.028 \\
 & \bridged \arc  & 0.765 & 0.545 & 0.655 & 81.4 & 0.079 & 0.143 \\
 & \bridged \ada  & 0.777 & 0.550 & 0.651 & 80.5 & 0.074 & \textbf{0.152} \\
 & \bridged \adavit & \textbf{0.796} & \textbf{0.590} & \textbf{0.649} & \textbf{75.2} & \textbf{0.082} & 0.146 \\
 & \bridged KPRPE & 0.795 & 0.578 & 0.652 & 77.4 & 0.075 & 0.140 \\
\midrule
\multicolumn{2}{l}{Arc2Face~\cite{arc2face} (ref.)} & 0.814 & 0.528 & 0.605 & 72.5 & 0.238 & 0.279 \\
\bottomrule
\end{tabular}}
\end{table}

% =====================================================================
\section{Naming a Face from Its Embedding}
\label{sec:naming}
% =====================================================================
Typically, naming an individual based on their face image needs an enrolled gallery: a probe embedding can then be compared against
stored embeddings of known people. However, when embeddings are aligned, names can be assigned to them even in the absence of an enrolled and labeled face gallery. Because a foundation model associates faces with names, an aligned
embedding can be matched against the foundation model's text embeddings of candidate
names, and the best-matching name can be selected. The gallery is simply a list of names, encoded once by the text encoder. Each name is encoded with four fixed prompts (``a photo of
\{name\}'', ``a portrait photo of \{name\}'', ``a face photo of the celebrity
\{name\}'', and the bare name), and the four text embeddings are averaged. The prompts
were fixed \textit{a priori} and are identical across all foundation models.

\noindent\textbf{Closed-set naming.}
We evaluate on CFP, whose 500 identities are named celebrities. Table~\ref{tab:naming} reports top-1, top-5 and top-10 naming
accuracy over the full 500-name vocabulary, where chance top-1 is 0.2\%. The \native
method names people at high accuracy (up to 88.7\% top-1 with CLIP). This is expected,
since these are public figures within the foundation model's pretraining data, and it
is precisely this knowledge that the alignment borrows. Every \bridged source names
held-out people far above chance while {\em Unaligned} and \emph{Random} sit exactly at chance. The IR-101 CNN embeddings reach 16--17\% top-1 through CLIP, more than 80 times
chance. The two ViT embeddings perform substantially better:
\adavit reaches 63.0\% top-1 and KPRPE 61.0\%, with 92.0\% and 91.6\% top-10 through
CLIP. Across the three foundation targets, the aligned ViT embeddings reach up to 73\%
of the native ceiling, and a ViT source is the best source for every target.

\begin{table}[tb]
\centering
\caption{Zero-shot naming and open-set watchlist screening on CFP (500-name vocabulary,
held-out identities). Closed-set: top-$k$ accuracy; open-set: DIR at fixed FAR and
rejection AUROC. Mean\,$\pm$\,std over five identity-disjoint splits.}
\label{tab:naming}
\setlength{\tabcolsep}{4pt}
\footnotesize
\resizebox{0.9\linewidth}{!}{%
\begin{tabular}{llccc|ccc}
\toprule
& & \multicolumn{3}{c|}{Closed-set} & \multicolumn{3}{c}{Open-set watchlist} \\
Foundation & Source & Top-1 & Top-5 & Top-10 & DIR@1\% & DIR@10\% & AUROC \\
\midrule
\multirow{7}{*}{CLIP}
& \native (ceiling) & 88.7\,{\tiny$\pm$0.8} & 95.2\,{\tiny$\pm$1.0} & 96.0\,{\tiny$\pm$0.9} & 55.5\,{\tiny$\pm$12.6} & 83.7\,{\tiny$\pm$2.5} & 0.941\,{\tiny$\pm$0.013} \\
\cmidrule(lr){2-8}
 & random transform.\ & 0.1\,{\tiny$\pm$0.1} & 0.7\,{\tiny$\pm$0.1} & 1.6\,{\tiny$\pm$0.2} & 0.0\,{\tiny$\pm$0.1} & 0.0\,{\tiny$\pm$0.1} & 0.502\,{\tiny$\pm$0.024} \\
 & \unaligned & 0.1\,{\tiny$\pm$0.1} & 0.8\,{\tiny$\pm$0.2} & 1.7\,{\tiny$\pm$0.2} & 0.0\,{\tiny$\pm$0.1} & 0.1\,{\tiny$\pm$0.1} & 0.504\,{\tiny$\pm$0.020} \\
 & \bridged \arc & 16.7\,{\tiny$\pm$1.4} & 41.6\,{\tiny$\pm$2.1} & 54.1\,{\tiny$\pm$2.2} & 6.2\,{\tiny$\pm$1.6} & 20.1\,{\tiny$\pm$2.8} & 0.659\,{\tiny$\pm$0.017} \\
 & \bridged \ada & 17.0\,{\tiny$\pm$0.7} & 39.9\,{\tiny$\pm$0.6} & 52.3\,{\tiny$\pm$1.1} & 5.5\,{\tiny$\pm$0.5} & 19.4\,{\tiny$\pm$1.6} & 0.649\,{\tiny$\pm$0.037} \\
 & \bridged \adavit & \textbf{63.0}\,{\tiny$\pm$1.4} & \textbf{87.2}\,{\tiny$\pm$1.2} & \textbf{92.0}\,{\tiny$\pm$0.8} & \textbf{25.9}\,{\tiny$\pm$7.7} & \textbf{61.5}\,{\tiny$\pm$3.9} & \textbf{0.863}\,{\tiny$\pm$0.020} \\
 & \bridged KPRPE & 61.0\,{\tiny$\pm$1.3} & 86.7\,{\tiny$\pm$1.1} & 91.6\,{\tiny$\pm$0.8} & 24.6\,{\tiny$\pm$5.0} & 57.1\,{\tiny$\pm$3.6} & 0.844\,{\tiny$\pm$0.019} \\
\midrule
\multirow{7}{*}{MetaCLIP}
& \native (ceiling) & 70.1\,{\tiny$\pm$2.1} & 87.7\,{\tiny$\pm$2.0} & 91.6\,{\tiny$\pm$1.6} & 35.5\,{\tiny$\pm$5.2} & 64.9\,{\tiny$\pm$2.4} & 0.860\,{\tiny$\pm$0.016} \\
\cmidrule(lr){2-8}
 & random transform.\ & 0.2\,{\tiny$\pm$0.1} & 0.9\,{\tiny$\pm$0.3} & 1.9\,{\tiny$\pm$0.4} & 0.0\,{\tiny$\pm$0.0} & 0.1\,{\tiny$\pm$0.1} & 0.488\,{\tiny$\pm$0.013} \\
 & \unaligned & 0.2\,{\tiny$\pm$0.1} & 0.9\,{\tiny$\pm$0.3} & 1.7\,{\tiny$\pm$0.3} & 0.0\,{\tiny$\pm$0.0} & 0.0\,{\tiny$\pm$0.0} & 0.482\,{\tiny$\pm$0.030} \\
 & \bridged \arc & 14.6\,{\tiny$\pm$0.7} & 37.1\,{\tiny$\pm$1.4} & 49.7\,{\tiny$\pm$1.8} & 4.7\,{\tiny$\pm$0.8} & 15.7\,{\tiny$\pm$1.2} & 0.631\,{\tiny$\pm$0.021} \\
 & \bridged \ada & 16.1\,{\tiny$\pm$0.4} & 37.0\,{\tiny$\pm$1.0} & 49.3\,{\tiny$\pm$1.1} & 7.0\,{\tiny$\pm$1.4} & 18.0\,{\tiny$\pm$1.8} & 0.613\,{\tiny$\pm$0.030} \\
 & \bridged \adavit & \textbf{50.9}\,{\tiny$\pm$1.6} & \textbf{78.8}\,{\tiny$\pm$2.5} & \textbf{86.5}\,{\tiny$\pm$1.7} & \textbf{20.9}\,{\tiny$\pm$5.2} & \textbf{47.1}\,{\tiny$\pm$2.6} & \textbf{0.802}\,{\tiny$\pm$0.012} \\
 & \bridged KPRPE & 48.6\,{\tiny$\pm$2.0} & 77.5\,{\tiny$\pm$1.5} & 85.6\,{\tiny$\pm$1.4} & 16.9\,{\tiny$\pm$3.9} & 45.0\,{\tiny$\pm$3.4} & 0.785\,{\tiny$\pm$0.020} \\
\midrule
\multirow{7}{*}{SigLIP}
& \native (ceiling) & 47.0\,{\tiny$\pm$1.6} & 69.1\,{\tiny$\pm$1.6} & 76.3\,{\tiny$\pm$1.0} & 21.0\,{\tiny$\pm$3.7} & 39.8\,{\tiny$\pm$2.3} & 0.749\,{\tiny$\pm$0.014} \\
\cmidrule(lr){2-8}
 & random transform.\ & 0.4\,{\tiny$\pm$0.4} & 0.9\,{\tiny$\pm$0.6} & 1.9\,{\tiny$\pm$0.8} & 0.1\,{\tiny$\pm$0.2} & 0.3\,{\tiny$\pm$0.5} & 0.491\,{\tiny$\pm$0.018} \\
 & \unaligned & 0.2\,{\tiny$\pm$0.1} & 0.9\,{\tiny$\pm$0.4} & 2.0\,{\tiny$\pm$0.6} & 0.0\,{\tiny$\pm$0.0} & 0.1\,{\tiny$\pm$0.1} & 0.486\,{\tiny$\pm$0.011} \\
 & \bridged \arc & 8.4\,{\tiny$\pm$1.0} & 22.7\,{\tiny$\pm$1.3} & 33.7\,{\tiny$\pm$2.0} & 2.2\,{\tiny$\pm$0.5} & 7.8\,{\tiny$\pm$0.6} & 0.581\,{\tiny$\pm$0.018} \\
 & \bridged \ada & 8.6\,{\tiny$\pm$0.8} & 24.9\,{\tiny$\pm$1.2} & 35.8\,{\tiny$\pm$1.8} & 2.4\,{\tiny$\pm$1.0} & 8.0\,{\tiny$\pm$1.4} & 0.565\,{\tiny$\pm$0.042} \\
 & \bridged \adavit & \textbf{26.1}\,{\tiny$\pm$2.0} & \textbf{51.6}\,{\tiny$\pm$1.7} & \textbf{63.1}\,{\tiny$\pm$1.4} & 5.5\,{\tiny$\pm$2.1} & \textbf{22.2}\,{\tiny$\pm$1.1} & \textbf{0.659}\,{\tiny$\pm$0.010} \\
 & \bridged KPRPE & 24.8\,{\tiny$\pm$2.1} & 49.1\,{\tiny$\pm$2.1} & 60.9\,{\tiny$\pm$1.7} & \textbf{6.9}\,{\tiny$\pm$2.9} & 21.1\,{\tiny$\pm$3.0} & 0.647\,{\tiny$\pm$0.019} \\
\bottomrule
\end{tabular}}
\end{table}

To check that this reflects the backbone rather than the training recipe, we additionally align an IR-101 CNN trained with the same AdaFace objective on
the same WebFace4M corpus~\cite{webface260m} as \adavit, so that architecture is the only difference between
the two. This model reaches 24.7\%, 21.5\% and 12.1\% top-1 through CLIP, MetaCLIP and
SigLIP,
against 63.0\%, 50.9\% and 26.1\% for \adavit: exchanging the convolutional backbone for a transformer, with the objective and
training data fixed, more than doubles the naming accuracy. All three foundation targets are themselves ViTs. A plausible hypothesis is therefore
that ViT-to-ViT alignment preserves identity directions that CNN-to-ViT alignment
attenuates, consistent with the model-family effects reported for cross-model
recognition~\cite{compat}. Naming is a stronger form of identity transfer than generation: enough identity
survives the linear transformation to select the correct person, by name, from 500
candidates, using only the embedding.

\begin{figure}[tb]
\centering
\includegraphics[width=0.8\linewidth]{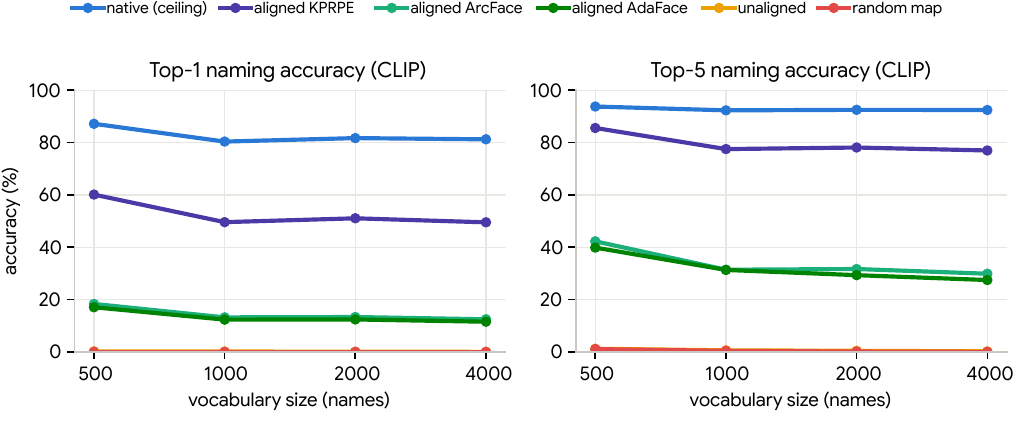}
\caption{Naming accuracy vs.\ vocabulary size (CLIP), 500 CFP names extended with FaceScrub
and synthetic distractors to 4{,}000. Aligned naming degrades in parallel with the native
ceiling; {\em Unaligned} and \emph{Random} stay at chance.}
\label{fig:namingabl}
\end{figure}

\noindent\textbf{Vocabulary size.}
We grow the vocabulary from 500 to 4{,}000 names by adding real celebrity distractors
(FaceScrub~\cite{facescrub}, with CFP overlaps removed) and then synthetic names formed
by randomly pairing first and last names from public name lists
(Fig.~\ref{fig:namingabl}). Accuracy degrades gradually and in parallel with the
\native ceiling. Through CLIP, aligned KPRPE falls from 61.0\% to 49.5\% (top-1 rank accuracy) and
aligned \arc from 16.7\% to 10.8\% as the vocabulary grows $8\times$. Over the same range, chance top-1 rank accuracy also falls from 0.2\% to 0.025\%.

\begin{figure}[tb]
\centering
\includegraphics[width=0.75\linewidth]{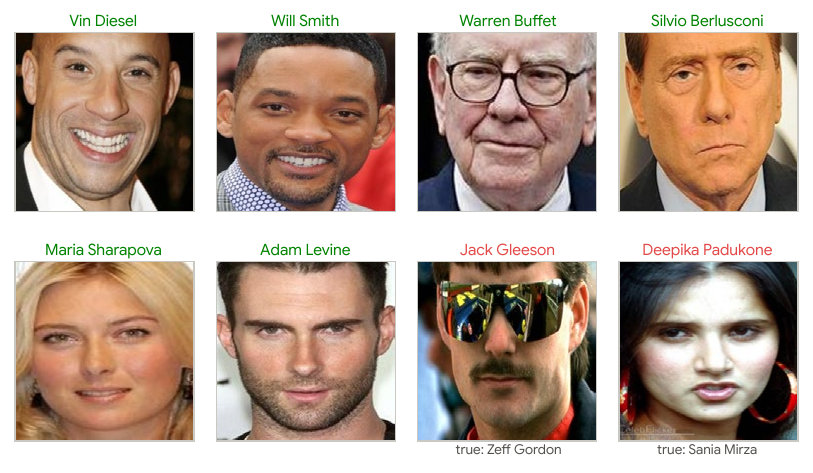}
\caption{Zero-shot naming (KPRPE $\to$ CLIP): held-out probes with the top predicted name
(green: correct, red: incorrect).}
\label{fig:naming}
\end{figure}

\noindent\textbf{Open-set watchlist screening.}
We then relax the closed-set assumption. The text vocabulary covers only half of the
test identities, forming a watchlist, and every test image is used as a probe. A probe is \emph{mated} if its identity is on the watchlist and \emph{non-mated} otherwise. A mated probe should be named, and a non-mated probe rejected by thresholding the top name score. We report two standard open-set metrics. (a) DIR@FAR$x$\%: the detection and identification rate, is the fraction of mated probes that are both accepted \emph{and} correctly named, at a threshold set so that only $x$\% of non-mated probes are falsely accepted. (b) The AUROC treats the top score as a detector separating mated from non-mated probes (0.5 = chance, 1 = perfect rejection). Table~\ref{tab:naming} reports both. Aligned embeddings perform well above chance but below the native ceiling. The gap is
wider here than for closed-set naming: the alignment preserves the rank of the correct
name better than it preserves the calibrated score margin needed to reject a
non-mated probe. Figure~\ref{fig:naming} shows named examples from held-out identities.

\noindent\textbf{Naming and web exposure.}
Naming succeeds because the foundation model has already seen the person, so accuracy
should depend on how often that person appears in web data. Using the number of images
per identity in LFW~\cite{lfw} as an exposure proxy, we bin the held-out probes and measure top-1
naming accuracy per bin (Table~\ref{tab:exposure}). Accuracy rises monotonically with
exposure: aligned KPRPE$\to$CLIP climbs from $11.6\%$ for identities with a single image
to $95.2\%$ for the most-photographed ones, tracking the native ceiling ($16.0\%$ to
$87.1\%$). Naming therefore reflects a person's representation in the pretraining data
rather than a fixed per-model limit.

\noindent\textbf{Cross-dataset transfer.}
All results so far estimate the map and evaluate it on the same dataset. To test whether
the alignment survives a change in data distribution, we fit the map on one dataset and
apply it, unchanged, to another: CFP$\to$LFW for naming and CelebA$\to$UTK for retrieval
(the \emph{cross} setting). We compare this against the standard \emph{within} setting,
where the map is fit on the target dataset's own training split. Both are evaluated on the
same target test set (Table~\ref{tab:cross}). The \native ceiling and the \unaligned and \emph{Random} floors
use no fitted map and are identical in both settings, so they are listed once. The ViT
sources transfer with little loss for naming (KPRPE
$25.4$ to $25.3$ top-1) and a moderate loss for retrieval ($0.761$ to $0.610$ mAP, still
well above the $0.28$ random floor), while the CNN sources degrade more in both tasks.
The larger retrieval drop is consistent with the wider domain gap between CelebA and
UTKFace than between CFP and LFW. A similar ordering, with ViT sources above CNN sources,
holds for the other two foundation targets.

\begin{table}[tb]
\centering
\begin{minipage}[t]{0.62\linewidth}
\centering
\caption{Cross-dataset transfer (CLIP). \emph{within}: map fit and evaluated on the target
dataset; \emph{cross}: map fit on a source dataset and evaluated on the target.}
\label{tab:cross}
\setlength{\tabcolsep}{4pt}
\renewcommand{\arraystretch}{0.9}
\scriptsize
\begin{tabular}{lcc|cc}
\toprule
& \multicolumn{2}{c|}{Naming top-1 (\%)} & \multicolumn{2}{c}{Retrieval mAP} \\
Source & within & cross & within & cross \\
\midrule
\native (ceiling)  & \multicolumn{2}{c|}{34.0} & \multicolumn{2}{c}{0.768} \\
random transform.\ & \multicolumn{2}{c|}{0.0}  & \multicolumn{2}{c}{0.279} \\
\unaligned         & \multicolumn{2}{c|}{0.1}  & \multicolumn{2}{c}{0.279} \\
\midrule
\bridged \arc    & 12.4 & 4.9  & 0.734 & 0.365 \\
\bridged \ada    & 12.3 & 6.0  & 0.723 & 0.415 \\
\bridged \adavit & 27.8 & 25.5 & 0.765 & 0.583 \\
\bridged KPRPE   & 25.4 & 25.3 & 0.761 & 0.610 \\
\bottomrule
\end{tabular}
\end{minipage}\hfill
\begin{minipage}[t]{0.35\linewidth}
\centering
\caption{Naming top-1 (\%) vs.\ web exposure: LFW probes binned by images per identity
(KPRPE$\to$CLIP).}
\label{tab:exposure}
\setlength{\tabcolsep}{4pt}
\renewcommand{\arraystretch}{0.9}
\scriptsize
\begin{tabular}{lcc}
\toprule
Images/id & \native & \bridged \\
\midrule
1      & 16.0 & 11.6 \\
2--3   & 28.7 & 22.1 \\
4--10  & 39.4 & 28.9 \\
11--50 & 51.3 & 34.4 \\
51+    & 87.1 & 95.2 \\
\bottomrule
\end{tabular}
\end{minipage}
\end{table}

% =====================================================================
\section{Discussion}
\label{sec:discussion}
% =====================================================================
\noindent\textbf{What transfers.}
The three tasks show that different kinds of information transfer to different
degrees. Coarse semantics and soft-biometric attributes transfer most readily, at close
to the native ceiling (Table~\ref{tab:text}). Appearance transfers well enough to drive
a generator, but stops short of fine identity; this is a limit of the
appearance-oriented decoders rather than of the alignment (Table~\ref{tab:gen}).
Identity transfers partially: enough to rank the correct name highly in a large
vocabulary, but not enough to produce the calibrated scores that open-set rejection
requires (Table~\ref{tab:naming}). Throughout, the alignment itself is a fixed linear
function of the embedding: it adds no information about the person, and information
that the source model never encoded cannot be recovered.

\noindent\textbf{What different embeddings expose.}
Because the probes are external and identical across models, they also permit a
comparison of the four FR models. For attributes and appearance, the four embeddings are nearly interchangeable. For
identity, they are not: the two ViT embeddings are named correctly
at $3.0$ to $3.7{\times}$ the rate of the two CNN embeddings, for every foundation
target, and the matched-pair comparison of Sec.~\ref{sec:naming} points to the backbone
rather than the training recipe. Four source models are too few to support a general
claim about architectures. Within this set, however, how much identity an embedding
exposed was governed by its backbone, and the same probes can be applied to any FR
model under consideration for deployment.

\noindent\textbf{Privacy implications.}
With only public models and one linear fit, a stored template becomes searchable,
renderable, and in some cases nameable. Unlike classical template
inversion~\cite{mai2019reconstruction,shahreza2024adapter,bhatta2025deep}, no attack
model needs to be trained against the specific FR model. Template-protection schemes
should therefore be evaluated against this channel as well. The same framework also
suggests a defense objective: a protection transform defeats this channel if it
destroys linear compatibility with public foundation spaces while preserving matching
accuracy. Designing and evaluating such alignment-resistant transforms is a
direction for future work.

\noindent\textbf{Limitations.}
The capabilities described in this paper are bounded by the models that provide them.
First, what the alignment exposes is limited by the expressivity of the target
foundation space: rendering recovers a likeness rather than recognition-grade identity
(Sec.~\ref{sec:generation}), and naming extends only to people represented in the
pretraining data, inheriting its demographic and coverage biases. Second, we deliberately keep the transformation linear,
since this is what makes it label-free, model-agnostic, and fast to fit. The cost is
that it can only expose information that is already linearly accessible in the
embedding. A more expressive, non-linear alignment might carry finer identity structure
across, but it moves back toward training a dedicated per-model network, which this
work deliberately avoids. Evaluating that trade-off is left to future work.

% =====================================================================
\section{Conclusion}
\label{sec:conclusion}
% =====================================================================
In this work, we investigated what capabilities become available when face embeddings
are made interoperable with foundation models. Our results show that such embeddings
can be unmasked by exploiting the linear alignment between the two families of models.
A single linear transformation, estimated once from paired FR and foundation-model
embeddings, is all that is required. Every downstream capability then follows without
labels, fine-tuning, or specialized networks. A face embedding can be searched with
free-form text, decoded into a realistic face image, and matched to a person's name,
while unaligned and randomly transformed embeddings perform at chance on every task.
Existing FR systems can thus inherit the semantic, generative, and open-vocabulary
capabilities of web-scale foundation models at negligible cost. At the same time, our
findings broaden the threat model for biometric embeddings: a representation stored
solely for matching may also disclose semantic attributes, visual appearance, and
name-level identity cues once aligned with publicly available foundation models.
Defending against this channel calls for template-protection schemes that resist
alignment itself~\cite{hahn2026handbook}.

\FloatBarrier
{\small
\bibliographystyle{splncs04}
\bibliography{main}
}
\end{document}